\documentclass[10pt,twocolumn,letterpaper]{article}

\usepackage{iccv}      

\definecolor{iccvblue}{rgb}{0.21,0.49,0.74}
\usepackage[pagebackref,breaklinks,colorlinks,allcolors=iccvblue]{hyperref}

\usepackage{multirow}
\usepackage[table,xcdraw]{xcolor}
\usepackage{colortbl}
\usepackage{rotating} 

\usepackage{pifont}
\usepackage{graphicx}    
\usepackage{capt-of}     

\usepackage{cuted}
\usepackage{capt-of}   
\usepackage{subcaption}

\def\paperID{1757073209586} 
\def\confName{ICCV}
\def\confYear{2025}

\title{TY-RIST: Tactical YOLO Tricks for Real-time Infrared Small Target Detection}

\author{
Abdulkarim Atrash$^{1}$ \quad Omar Moured$^{2}$\thanks{Corresponding author: \texttt{moured.omar@gmail.com}} \quad Yufan Chen$^{2}$ \\
Jiaming Zhang$^{2}$ \quad Seyda Ertekin$^{1}$ \quad Ömür Uğur$^{1}$ \\
\\
$^{1}$Middle East Technical University \\
{\tt\small \{atrash.abdulkarim, sertekin, ougur\}@metu.edu.tr} \\
$^{2}$Karlsruhe Institute of Technology \\
{\tt\small \{name.surname\}@kit.edu}
}

\begin{document}
\maketitle

\definecolor{lightgreen}{HTML}{CCFFCC}
\definecolor{lightred}{HTML}{FFCCCC}
\begin{strip}
  \vspace{-3em}
  \centering
  \includegraphics[width=\textwidth]{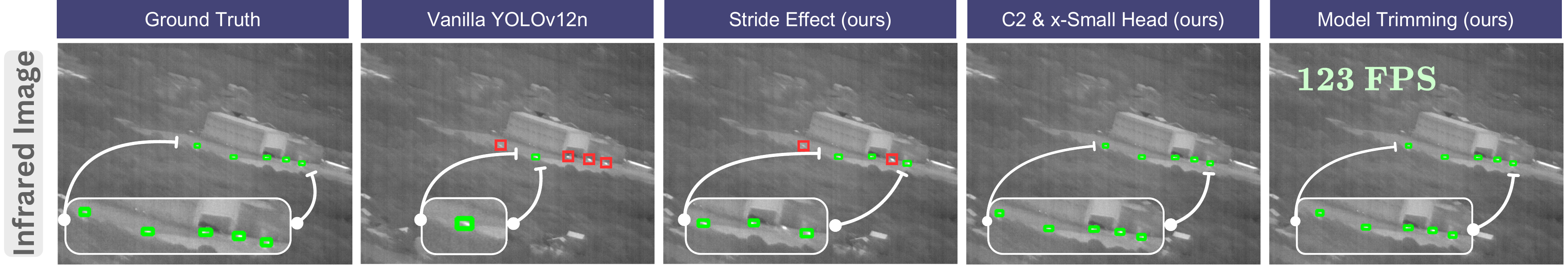}
    \captionof{figure}{Qualitative comparison of our model TY-RIST with the baseline, showing the effects of stride reduction, the higher-resolution feature map ($C_{2}$) with its detection head ($P_{2}$), and model trimming. \colorbox{lightgreen}{Green} boxes denote true positives; \colorbox{lightred}{red} boxes denote false negatives.}
  \label{fig:teaser}
\end{strip}

\begin{abstract}
While critical for defense and surveillance, infrared small target detection (IRSTD) remains a challenging task due to: (1) target loss from minimal features, (2) false alarms in cluttered environments, (3) missed detections from low saliency, and (4) high computational costs. To address these, we propose TY-RIST, an optimized YOLOv12n architecture featuring: (1) a stride-aware backbone with fine-grained receptive fields, (2) a high-resolution detection head, (3) cascaded coordinate attention blocks, and (4) a branch pruning strategy that reduces computational cost up to $\sim$$25.5$\% while marginally enhancing performance and enabling real-time inference. Additionally, we incorporate the Normalized Gaussian Wasserstein Distance (NWD) to improve regression stability. Extensive experiments on four benchmarks and across $20$ different models demonstrate state-of-the-art performance, boosting mAP@$50$ by {+}$7.9$\%, Precision by {+}$3$\%, and Recall by {+}$10.2$\%, while running up to $\sim$$123$ FPS on a single GPU. Cross-dataset validation on a fifth dataset further confirms strong generalization capability. Further results and details are published at \url{www.github.com/moured/TY-RIST}.
\end{abstract}
    
\section{Introduction}
\label{sec:intro}

In the field of object detection, small targets are formally defined as objects occupying less than 0.5\% of the total image area, exhibiting weak contrast (typically below 0.15), and possessing a low signal-to-noise ratio (SNR) \cite{chapple1999target}. IRSTD focuses on detecting small and often moving targets embedded within cluttered and noisy infrared backgrounds. IRSTD holds significant importance for critical applications, including military reconnaissance \cite{tang2016novel}, traffic monitoring and management \cite{zhang2021review}, and maritime search and rescue operations \cite{teutsch2010classification}. Nevertheless, IRSTD remains a highly challenging task due to several inherent difficulties. First, the minimal size and weak signal strength of targets often lead to the loss of critical features, compromising reliable detection. Second, cluttered or textured backgrounds contribute to elevated false-alarm rates. Third, low target saliency and contrast frequently result in missed detections. Fourth, the high computational demands of advanced detection frameworks limited their practical deployment in real-time applications.

IRSTD methods are categorized based on feature utilization, problem formulation, and algorithmic approach. Regarding feature extraction, algorithms are divided into single-frame approaches (SIRST) \cite{zhu2023yolo}, which process spatial features from individual frames, and multi-frame approaches (MIRST) \cite{peng2025moving}, which exploit temporal features from video sequences to enhance detection performance at increased computational cost. From a formulation perspective, SIRST implementations may adopt either a detection-based paradigm \cite{zhu2023yolo} or a segmentation-based approach \cite{li2022dense}. MIRST, on the other hand, has only been formulated as a detection problem so far due to the absence of a suitable multi-frame segmentation-based dataset. Algorithmically, while classical techniques such as filtering \cite{bae2011small} and local contrast enhancement \cite{han2020infrared} demonstrate computational efficiency, their dependence on expert-driven parameter tuning limits their generalization capability.

\begin{figure}[t]
  \centering
   \includegraphics[width=1.0\linewidth]{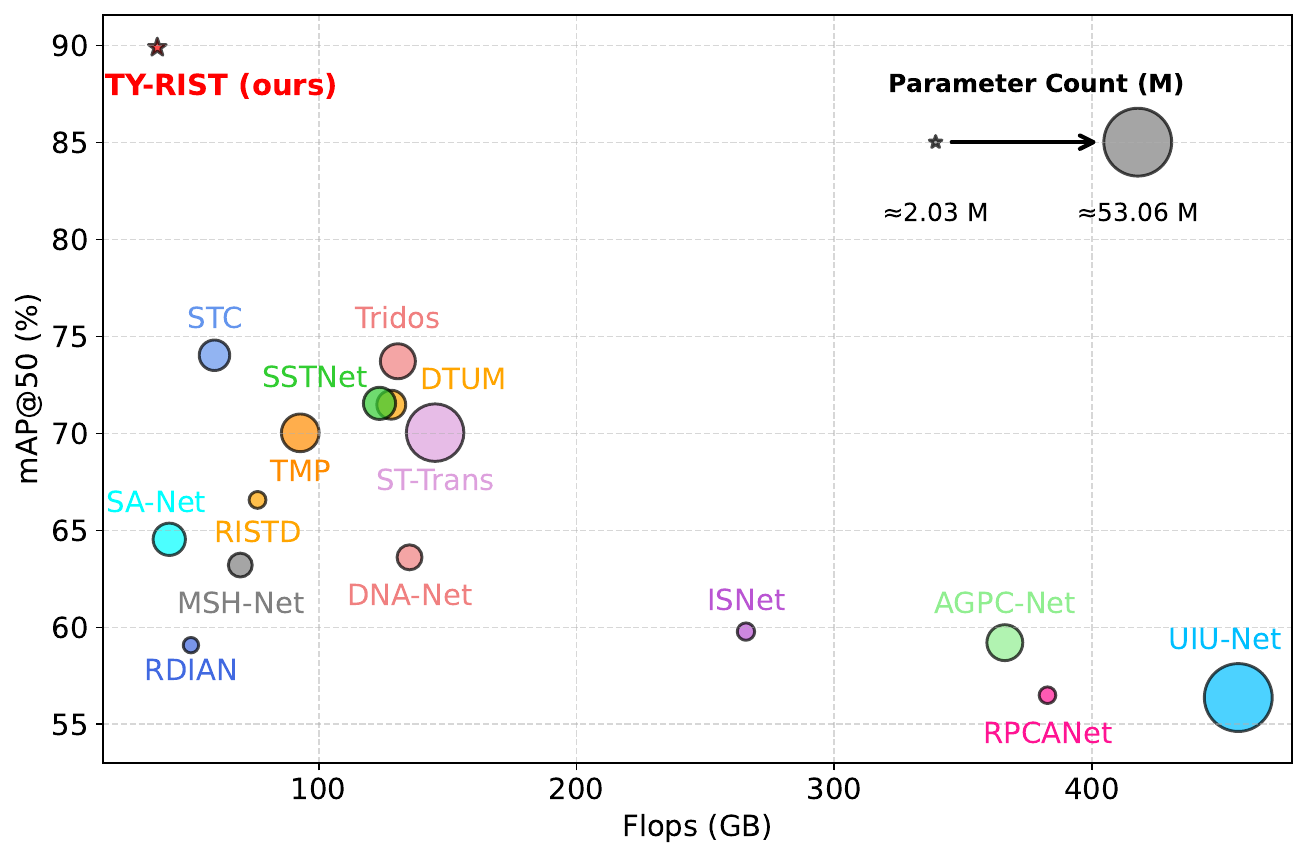}

   \caption{Detection performance vs.\ GFLOPs on the IRDST dataset. Circle size indicates model parameter count, and our TY-RIST model is denoted by \textcolor{red}{$\bigstar$}.}
   \label{fig:map vs gflops IRDST}
\end{figure}

In contrast, deep learning-based approaches have recently achieved notable success in IRSTD, offering high accuracy and strong generalization in complex infrared scenes. Among these, YOLO-based detectors \cite{zhu2023yolo,li2023yolosr} have gained significant attention due to their ability to balance detection performance with real-time inference efficiency effectively. YOLOv12 \cite{tian2025yolov12}, the most recent advancement in the YOLO family, has introduced a novel attention mechanism, Area Attention, which enhances feature representation and surpasses traditional convolutional architectures while maintaining competitive inference speed. 

We present \textbf{TY-RIST}, Tactical YOLO Tricks for Real-time Infrared Small Target Detection, a unified framework based on the YOLOv12n baseline that addresses the aforementioned challenges. First, we introduce a stride-aware convolutional backbone to construct a fine-grained receptive field for improved spatial localization. Second, we add a high-resolution feature map with a dedicated tiny-object detection head to suppress false alarms. Third, we integrate cascaded Coordinate Attention (CA) \cite{hou2021coordinate} blocks on the newly added detection head to reduce missed detections. Fourth, we replace the classical Complete Intersection over Union (CIoU) \cite{zheng2021enhancing} loss with the NWD \cite{wang2021normalized} to improve regression stability and ease convergence, addressing the sensitivity of bounding box regression to infrared small targets. Fifth, we optimize the architecture by pruning redundant branches, achieving up to $\sim$\textbf{$25.5$\%} reduction in GFLOPs and up to $\sim$\textbf{$25.6$\%} in the number of parameters while offering incremental improvement in performance and running at real-time speed. Figure~\ref{fig:teaser} qualitatively illustrates the impact of some of the proposed experiments on the model’s performance, demonstrating reduced missed detections and real-time inference capability.

We evaluated our model on two multi-frame (ITSDT-15k~\cite{zhu2024tmp} and IRDST~\cite{sun2023receptive}), and two single-frame benchmarks (NUAA-SIRST \cite{dai2021asymmetric} and NUDT-SIRST \cite{li2022dense}), achieving improvements up to \textbf{$7.9$\%} in mAP@$50$,  \textbf{$3$\%} in Precision, and \textbf{$10.2$\%} in Recall over the baseline on ITSDT-15k benchmark. Furthermore, our model outperforms \textbf{$14$} SIRST and \textbf{$6$} MIRST state-of-the-art (SOTA) algorithms on the four benchmarks (results on the IRDST benchmark are shown in Figure~\ref{fig:map vs gflops IRDST}), while maintaining a real-time inference speed up to $\sim$\textbf{$123$} FPS on a single NVIDIA RTX3080 Ti GPU. Finally, a cross-dataset validation on the unseen IRDST-1k \cite{zhang2022isnet} benchmark confirmed strong generalization capability. 

\definecolor{highlightbg}{HTML}{FFE6F2}

\begin{figure*}
    \centering
    \includegraphics[width=\linewidth]{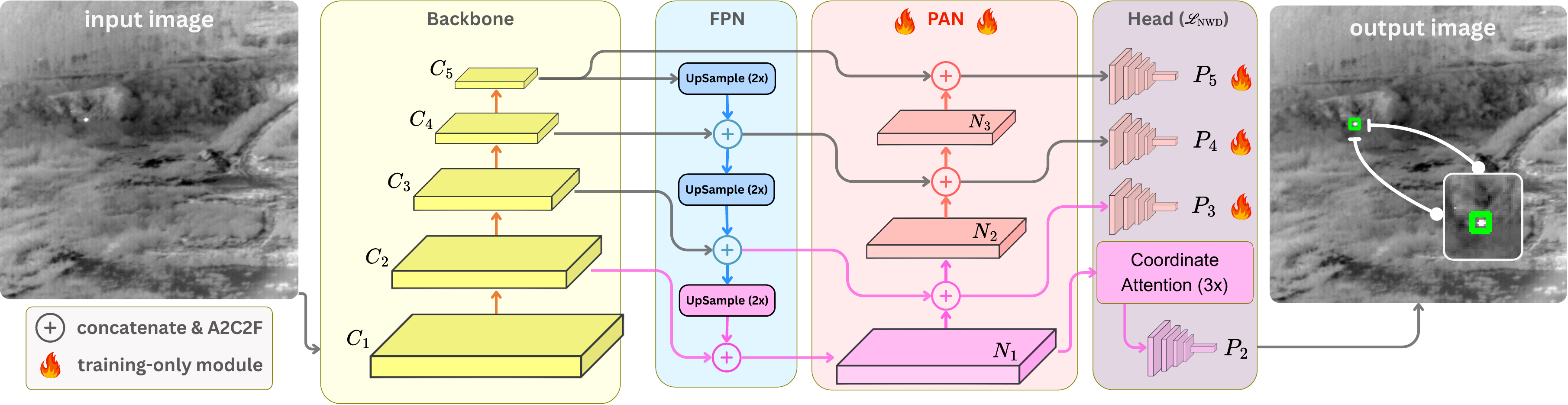}
    \caption{An overview of our TY-RIST, based on the YOLOv12n, which incorporates improved backbone, neck, and head modules. \colorbox{highlightbg}{Pink} color shows our added modules.}
    \label{fig: main figure}
\end{figure*}

\section{Related Work}
\label{sec:related}

\subsection{Data Driven SIRST Paradigm}
Learning-based SIRST approaches have become dominant in IRSTD by leveraging attention mechanisms, advanced feature modeling, and contextual reasoning. Channel and spatial attention modules \cite{dai2021asymmetric,li2022dense,zhang2022isnet,zhu2023sanet,zhang2023attention,xu2024hcf} enhance fine details, shape cues, and global-local correlations. Other methods integrate multi-scale and hybrid feature extraction \cite{hou2021ristdnet,sun2023receptive,wu2022uiu,yang2024eflnet,wu2024rpcanet} to better capture tiny targets. Recent work also emphasizes dataset and loss design, including negative sample augmentation \cite{lu2024sirst} and multi-scale heads with novel losses \cite{liu2024infrared}, showing that architectural and data-centric innovations are equally important.

\subsection{Data Driven MIRST Paradigm}
Multi-frame SIRST methods enhance detection by exploiting temporal features across sequences, which helps suppress false alarms and reinforce weak targets. ConvLSTM-based spatio-temporal fusion networks \cite{chen2024sstnet,zhu2024tmp,duan2024triple,li2023direction} capture motion cues, complementary features, and direction information to strengthen temporal consistency. Inspired by the success of transformers in vision \cite{dosovitskiy2020image}, recent works extend this to IRSTD by modeling frame-to-frame dependencies \cite{tong2024st} or jointly learning spatial, temporal, and channel correlations \cite{zhu2025spatial}. While transformer-based models achieve strong performance, their computational demands pose challenges for real-time deployment.

\section{Methodology}
\label{sec:methodology}

\subsection{Overall Architecture}

The overall architecture of the proposed framework is summarized in Figure~\ref{fig: main figure}. It is based on the YOLOv12n \cite{tian2025yolov12} architecture, with a series of improvements applied at each stage of the pipeline. This section presents a detailed explanation of each conducted experiment.

\subsubsection{Stride Effect}

IRSTD suffers from the limited spatial features of tiny objects. While increasing input resolution or applying super-resolution can alleviate this issue, such methods~\cite{hao2024infrared, li2023yolosr,ren2022infrared,yue2025yolo} often rely on two-stage pipelines that hinder real-time applicability. Motivated by this, we propose a novel approach that avoids enlarging the input image and instead focuses on enlarging the backbone’s feature maps by reducing the stride in the first CNN block from $2$ to $1$, thereby duplicating the produced feature maps throughout the entire model by a factor of two.

\begin{figure}[h]
    \centering
    \includegraphics[width=\linewidth]{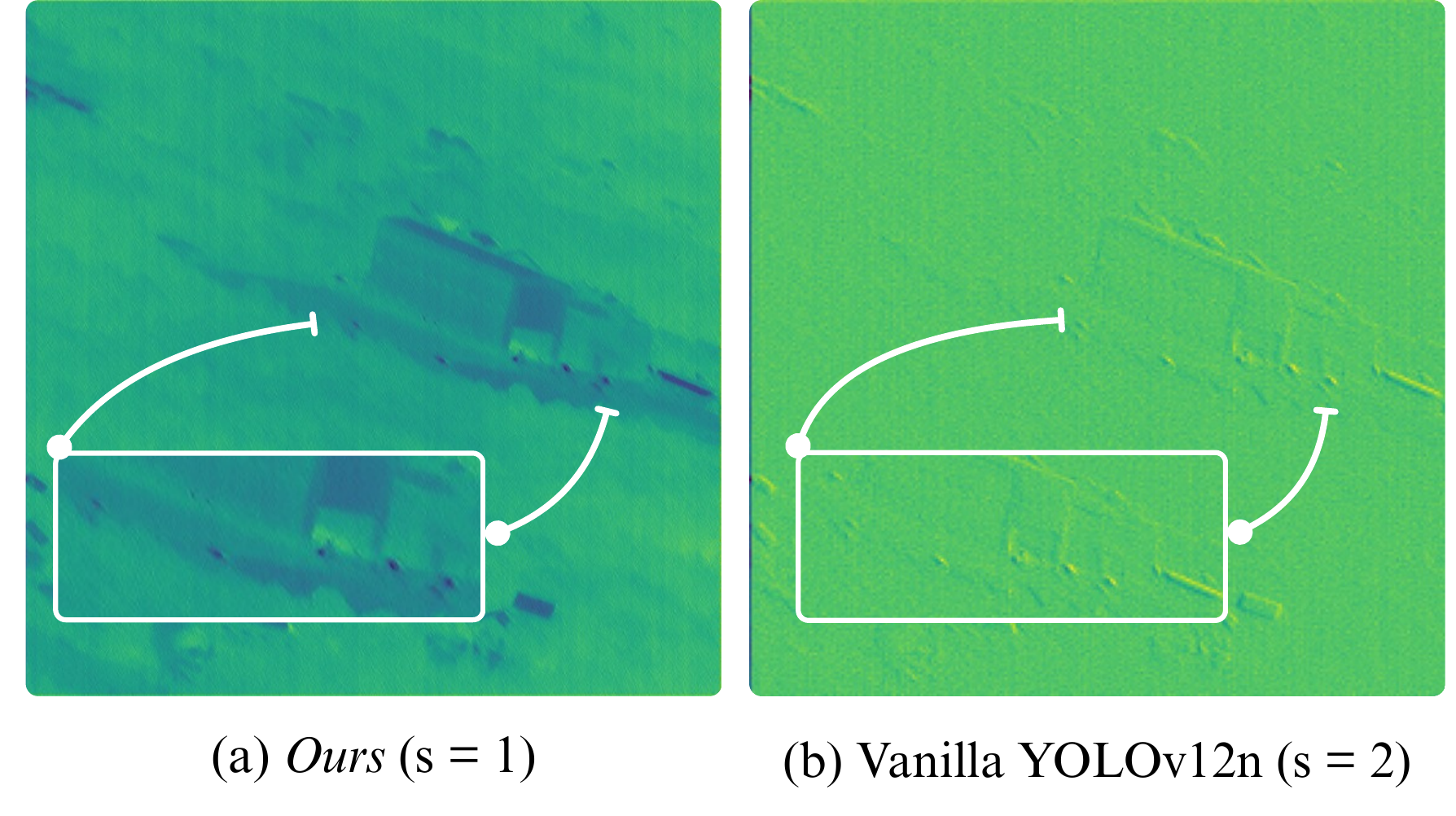}
    \caption{Comparison of the feature maps from filter $5$ of the first convolutional layer for the two models, with and without stride reduction. In (a), fine details are preserved, whereas in (b), the tiny targets appear blurred out.}
    \label{fig:filter5_comparison}
\end{figure}

Figure~\ref{fig:filter5_comparison} visualizes the feature maps obtained from two models with and without stride reduction. Reducing the stride preserves critical fine details that are propagated through the subsequent layers of the backbone network, whereas a stride of $2$ results in the loss of such critical features.

\subsubsection{Regression Loss via NWD Function}

\begin{figure}[htbp]
  \centering
   \includegraphics[width=1.0\linewidth]{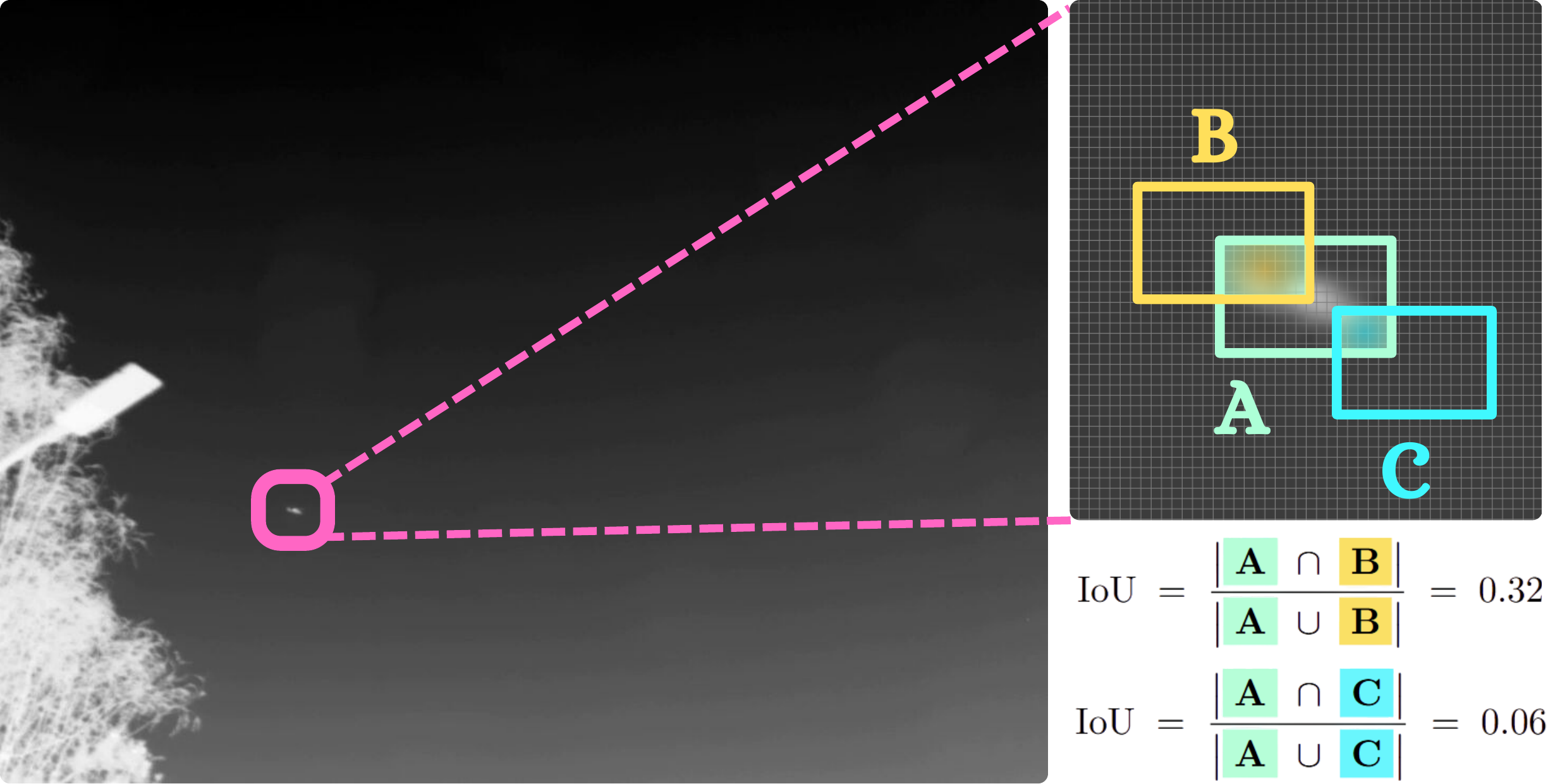}

   \caption{A case illustrating CIoU’s sensitivity to small objects.}
   \label{fig:nwd}
\end{figure}

Intersection over Union (IoU) based metrics, such as the CIoU function \cite{zheng2021enhancing}, are commonly used for bounding box regression in generic object detection. However, they are highly sensitive in the context of small target detection, where even minor positional deviations between predicted and ground-truth boxes can cause significant drops in IoU. For example, as shown in Figure~\ref{fig:nwd}, the IoU between the ground-truth bounding box A and predicted boxes B and C dropped sharply from $0.32$ to $0.06$ despite only small positional differences.

Following the related literature~\cite{li2025ipd, yang2024eflnet,zhou2022infrared}, the NWD function \cite{wang2021normalized} was adopted to replace the CIoU function, as it addresses the aforementioned issue by modeling bounding boxes as $2$D Gaussian distributions and measuring the similarity between them using the Wasserstein distance, making it insensitive to differences in object scale and robust to minimal or no overlap. The $2$D Wasserstein distance between two $2$D Gaussian distributions $\mu_1$ = $N(\textbf{m}_{\mathbf{1}} , \mathbf{\Sigma_{1}})$ and $\mu_{2} = N(\mathbf{m}_{\textbf{2}}, \mathbf{\Sigma_{2}})$ is defined in Equation~\ref{eqn:w2}:

\begin{equation}
W_2^2(\mu_{1}, \mu_{2}) = \|\mathbf{m}_1 - \mathbf{m}_2\|_2^2 +  \| \mathbf{\Sigma_{1}^{\frac{1}{2}}} - \mathbf{\Sigma_{2}^{\frac{1}{2}}} \|_{F}^{2} ,
\label{eqn:w2}
\end{equation}

\noindent where $\| . \|_{F}$ is the Frobenius norm, $\mathbf{m}$ is the mean vector, and $\mathbf{\Sigma}$ is the covariance matrix.  The distance between the
$2$ Gaussian distributions $\mathcal{N}_a$, $\mathcal{N}_b$ modeled by bounding boxes $A$ = ($cx_{a}$, $cy_{a}$, $w_{a}$, $h_{a}$) and $B$ = ($cx_{b}$, $cy_{b}$, $w_{b}$, $h_{b}$) can be written as Equation~\ref{simplified}:

\begin{equation}
\begin{split}
W_2^2(\mathcal{N}_a, \mathcal{N}_b) 
&= \left|\left|\left( 
    \left[ c x_a,\, c y_a,\, \tfrac{w_a}{2},\, \tfrac{h_a}{2} \right]^{\top} 
    \right.\right.\right. \\[-2pt]
&\quad\left.\left.\left.
    - \left[ c x_b,\, c y_b,\, \tfrac{w_b}{2},\, \tfrac{h_b}{2} \right]^{\top} 
\right)\right|\right|^2_{2},
\end{split}
\label{simplified}
\end{equation}

\noindent where $c_x$ and $c_y$ denote the coordinates of the bounding box center, and $w$ and $h$ denote its width and height, respectively. Normalizing it exponentially to a range of $0-1$ gives the NWD \cite{wang2021normalized} function in Equation~\ref{nwd}.

\begin{equation}
    NWD(\mathcal{N}_a, \mathcal{N}_b) = \exp \left( - \frac{\sqrt{W_2^2(\mathcal{N}_a, \mathcal{N}_b)}}{C} \right),
    \label{nwd}
\end{equation}

\noindent where $C$ is a dataset-dependent constant, treated as a hyperparameter and requiring fine-tuning.

\subsubsection{Higher Resolution Features \& x-Small Head}

IRSTD suffers from high false alarm rates, reflected in low precision. While YOLOv12n uses multi-scale heads (P$_3$, P$_4$, P$_5$) to detect small, medium, and large objects, its reliance on heavily downsampled feature maps causes loss of critical spatial details, making it ineffective for extra-small infrared targets. The main limitation of existing architectures is their exclusion of the shallow C$_2$ feature map, which, due to its high resolution, is crucial for detecting weak and tiny targets.

\begin{figure}[h]
    \centering
    \includegraphics[width=\linewidth]{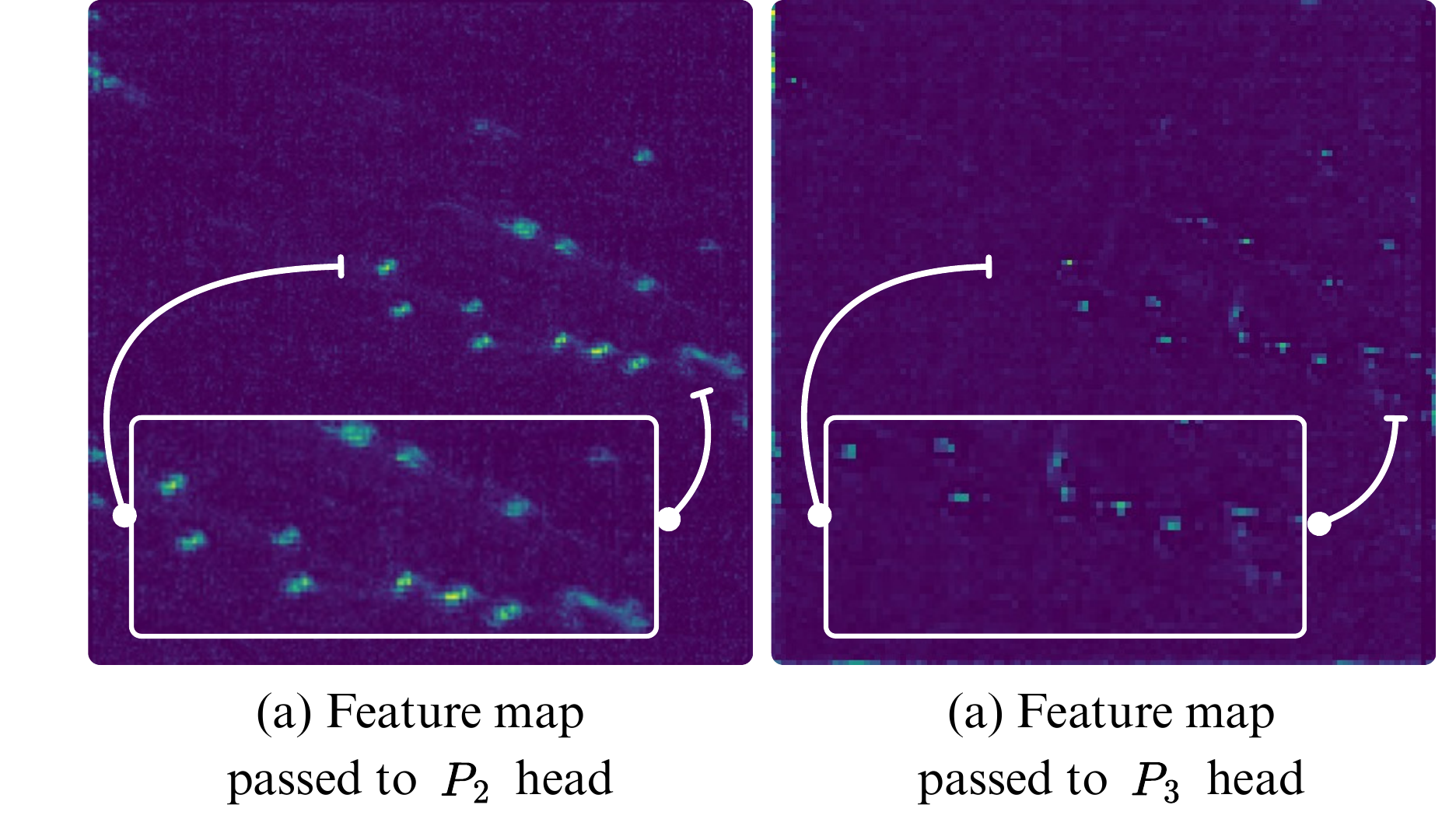}
    \caption{Comparison of the feature maps passing to $P_{2}$ and $P_{3}$. In (a), the target features are more clearly visible with higher resolution compared to (b).}
    \label{fig:P2 features}
\end{figure}

 \noindent Inspired by prior work~\cite{chu2025lmsfa,li2023yolosr,qi2025investigation}, we incorporate C$_2$ (shown in pink concatenation sign in Figure~\ref{fig: main figure}) alongside C$_3$, C$_4$, and C$_5$, modify the neck to produce a P$_2$ feature map, and extend the detection head with Head $2$ (shown in pink head module in Figure~\ref{fig: main figure}), dedicated to tiny target detection. Figure~\ref{fig:P2 features} illustrates the difference in the resolutions of the feature maps passed to the $P_2$ and $P_3$ heads, respectively. Feature maps extracted from $P_2$ head are of higher resolution and thus of richer meaning.

\subsubsection{The Addition of Coordinate Attention (CA) Blocks}

In IRSTD, missed detections (false negatives) reduce recall by failing to identify true targets. To address this, following the related literature~\cite{da2024infrared,li2023yolosr,shi2023infrared}, we incorporate Coordinate Attention~\cite{hou2021coordinate} (CA) on the highest resolution detection head branch ($P_2$ head, shown in pink block in Figure~\ref{fig: main figure}), which enables a network to understand not only what parts of an image are important but also where they are located. Traditional attention mechanisms, such as Convolutional Block Attention Module (CBAM)~\cite{woo2018cbam} and Squeeze and Excite (SE)~\cite{hu2018squeeze}, often emphasize important features while losing precise positional information due to global pooling over both spatial dimensions. Coordinate attention~\cite{hou2021coordinate}, on the other hand, addresses this limitation by decomposing spatial pooling into two one-dimensional operations: one along the horizontal direction and the other along the vertical direction. This allows the network to capture long-range dependencies in one direction while retaining location information in the other. 

\begin{figure} [h]
    \centering
    \includegraphics[width=\linewidth]{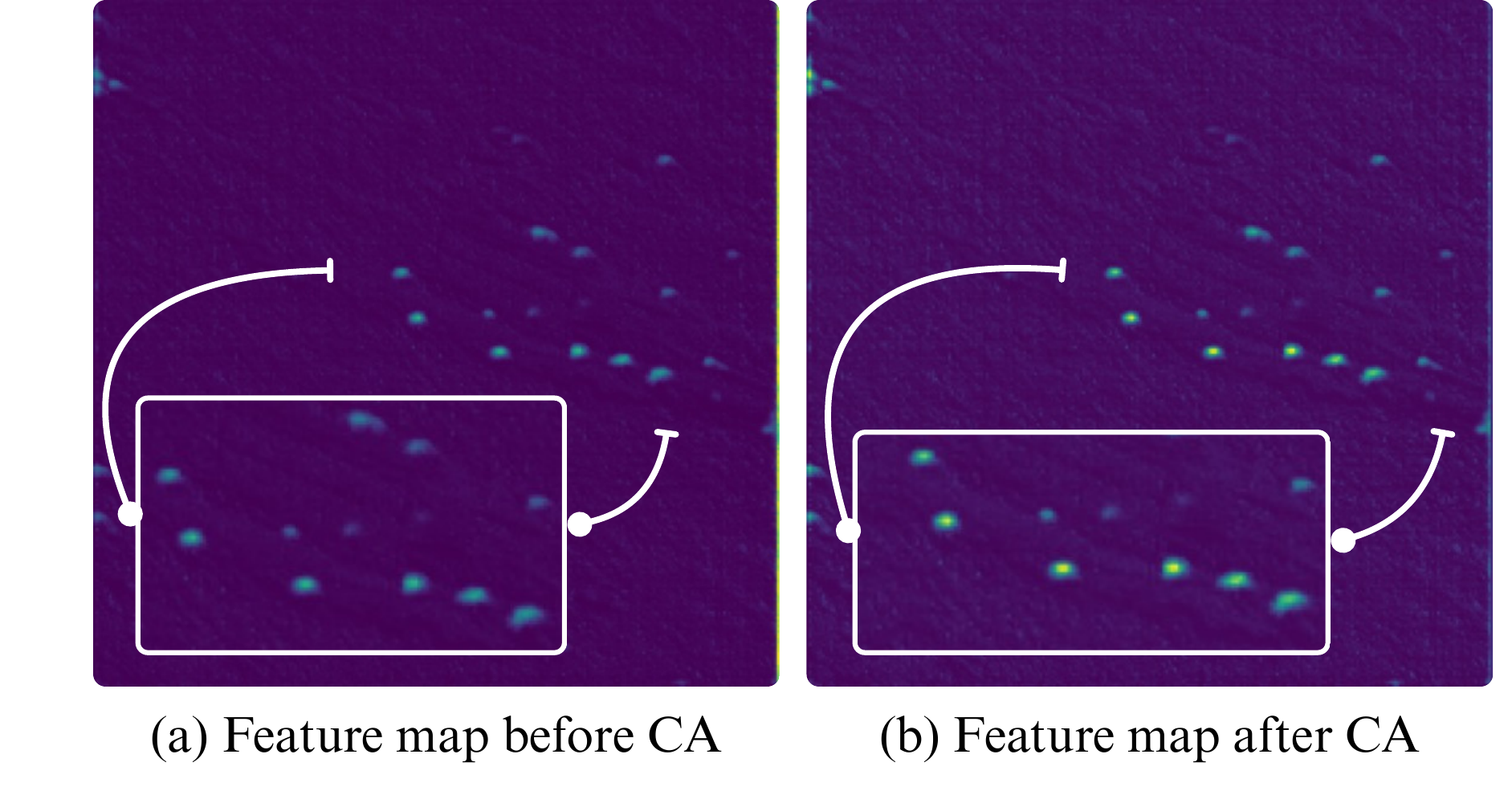}
    \caption{Comparison of the feature maps before and after passing through three CA blocks. In (a), the tiny target features appear faded and blurred, whereas (b) shows brighter, more prominent features resulting from the effect of the $3$ CA blocks.}
    \label{fig:ca features}
\end{figure}

Figure~\ref{fig:ca features} presents a visual comparison of the feature map $P2$ before and after applying the three coordinate attention blocks. The weak target features were enriched and more focused after applying the CA blocks.

\subsubsection{Model Optimization}

Real-time IRSTD requires lightweight algorithms that ensure fast, accurate performance on resource-constrained platforms, enabling timely responses in dynamic applications.
The original YOLOv12n model employs three detection heads for small, medium, and large objects, supplemented by an additional $P_{2}$ head specifically designed for extra-small objects. To assess the contribution of each head, and motivated by the related literature~\cite{wang2025ow,xiao2025fbrt}, we conducted a trimming experiment by systematically disabling one head at a time during inference (indicated by flame icons in Figure~\ref{fig: main figure}). On the ITSTD-15k benchmark~\cite{zhu2024tmp}, using only the $P_{2}$ head not only preserved performance but also slightly improved it by reducing both error propagation from other heads and model complexity, as evidenced by lower FLOPS and fewer parameters. This improvement arises because the $P_{2}$ head processes the highest-resolution feature map ($C_{2}$), enhanced by three CA blocks before prediction, making it ideally suited for IRSTD. However, on the NUAA-SIRST benchmark~\cite{dai2021asymmetric}, two heads ($P_{2}$ and $P_{3}$) were necessary for obtaining the optimal performance due to the presence of larger-sized objects that head $P_{2}$ failed to fully detect. To further investigate the errors introduced by other heads, we replaced the PAN network (fully on ITSTD-15 benchmark and partially on NUAA-SIRST benchmark to the production of N$_1$ feature map which is used to produce $P_3$ head, shown in Figure~\ref{fig: main figure}) with an identity matrix during inference, effectively eliminating feature aggregation. The performance remained comparable to using the related heads but with reduced complexity. This confirms that the PAN network's downsampling—necessary to distribute features to the unutilized heads degrades performance by potentially discarding critical target features. Consequently, removing (fully or partially) the PAN network had no adverse effect, which is consistent with our earlier findings on the drawbacks of downsampling.

\section{Experimental Settings}
\label{sec:dataset and metrics}

\subsection{Benchmark Datasets}

We conducted our experiments on five publicly available datasets with bounding box annotations: two sequence-based datasets (\textbf{IRDST} \cite{sun2023receptive}, and \textbf{ITSDT-15k} \cite{zhu2024tmp}) and three single-frame datasets (\textbf{NUAA-SIRST} \cite{dai2021asymmetric}, \textbf{NUDT-SIRST} \cite{li2022dense}, and \textbf{IRDST-1k} \cite{zhang2022isnet}). ITSDT-15k \cite{zhu2024tmp}, derived from the original $87$-sequence ITSDT dataset \cite{fu2022dataset}, contains challenging air-to-ground moving vehicle scenes with occlusion, blurring, and rotation. IRDST includes $85$ real and $317$ simulated ground-to-air sequences for flying target detection. For IRDST, we followed the training and validation splits defined by \cite{chen2024sstnet}. The single-frame datasets, originally annotated at the pixel level, encompass a variety of complex backgrounds such as clouds, cities, rivers, roads, seas, and fields. For our experiments, we utilized the bounding box annotations and dataset splits provided by \cite{yang2024eflnet}.

\subsection{Evaluation Metrics}
\label{metrics}
Following the common practice in solving IRSTD by detection paradigm, we evaluated performance using Precision (\%), Recall (\%), F1 score (\%), and Average Precision (\%) (e.g., mAP$50$). In addition, we report the number of model parameters in millions (M) and the computational cost in terms of floating point operations (FLOPS) measured in Giga.

\subsection{Implementation Details}

Our experiments consist of three parts. First, for multi-frame benchmarks (IRDST \cite{sun2023receptive} and ITSDT-15k \cite{zhu2024tmp}), we resized input images to $512\times512$ following \cite{chen2024sstnet}, initialized YOLOv12n with COCO \cite{lin2014microsoft} weights, and trained for $100$ epochs using AdamW \cite{loshchilov2017decoupled} with a learning rate of $0.0001$ and batch size of $4$. This setup was used for experiments involving stride reduction, replacing CIoU \cite{zheng2021enhancing} with NWD \cite{wang2021normalized}, and adding $P_{2}$ and x-small head modules. For the CA \cite{hou2021coordinate} experiment, we adopted a two-stage training strategy by freezing the backbone and neck, reinitializing the head with COCO weights, adding CA blocks only to the x-small head branch, and fine-tuning the added CA and head parts for $100$ epochs. Second, for single-frame benchmarks (NUAA-SIRST \cite{dai2021asymmetric}, NUDT-SIRST \cite{li2022dense}), we used the same settings except for increasing image resolution to $640\times640$, following \cite{yang2024eflnet}, and training for $200$ epochs. Finally, we conducted a cross-dataset validation experiment by combining NUAA-SIRST and NUDT-SIRST for training and validating on IRDST-1k \cite{zhang2022isnet}. Our training experiments were conducted on a cluster node equipped with a single NVIDIA A$40$ GPU with $45$ GB of memory, while inference experiments were conducted on a laptop with a single NVIDIA RTX $3080$ Ti GPU. We report results for a single training trial. Repeating experiments is advised for more reliable statistical analysis.

\definecolor{benchmark}{HTML}{F0F8FF}     
\definecolor{best}{HTML}{E8FFE8}          
\definecolor{delta_pos}{HTML}{006400}     
\definecolor{delta_neg}{HTML}{F44336}     
\definecolor{secondbest}{HTML}{FFEFD5}   

\begin{table*}[h]
  \centering
  \Large
  {\renewcommand{\arraystretch}{1.3}%
   \setlength{\tabcolsep}{4pt}%
   \resizebox{\textwidth}{!}{%
     \begin{tabular}{c c c *{2}{c} *{2}{c} c c}
       \toprule
       \cellcolor{benchmark}\textbf{Method}
         & \cellcolor{benchmark}\textbf{Year}
         & \cellcolor{benchmark}\textbf{Venue}
         & \multicolumn{2}{c}{\cellcolor{benchmark}\textbf{IRDST (\%)}}
         & \multicolumn{2}{c}{\cellcolor{benchmark}\textbf{ITSDT-15k (\%)}}
         & \cellcolor{benchmark}\textbf{Params (M) {\color{delta_pos}\(\downarrow\)}}
         & \cellcolor{benchmark}\textbf{Flops (G) {\color{delta_pos}\(\downarrow\)}} \\
       & & 
         & \textbf{$\mathrm{mAP}_{50}$} {\color{delta_pos}\(\uparrow\)}
         & \textbf{$F_{1}$}      {\color{delta_pos}\(\uparrow\)}
         & \textbf{$\mathrm{mAP}_{50}$} {\color{delta_pos}\(\uparrow\)}
         & \textbf{$F_{1}$}      {\color{delta_pos}\(\uparrow\)}
         &               &               \\
       \cmidrule(lr){4-5}\cmidrule(lr){6-7}
       \multicolumn{9}{l}{\textit{\textbf{Single‐frame Methods}}} \\
       HCFNet \cite{xu2024hcf}           & 2024 & ICME           & –     & –       & 57.54 & 76.20   & –     & –      \\
       
       AGPCNet \cite{zhang2023attention}& 2023 & IEEE TAES      & 59.21 & 77.44   & 67.27 & 82.16   & 14.88 & 366.15 \\
       
       DNANet \cite{li2022dense}        & 2023 & IEEE TIP       & 63.61 & 80.11   & 70.46 & \cellcolor{best} \textbf{84.46}   &  7.20 & 135.24 \\
       
       RISTD \cite{hou2021ristdnet}     & 2022 & IEEE GRSL      & 66.57 & 82.08   & 60.47 & 77.93   &  3.28 &  76.28 \\
       
       ISNet \cite{zhang2022isnet}      & 2022 & CVPR           & 59.78 & 77.58   & 62.29 & 79.18   &  3.49 & 265.73 \\
       
       RDIAN \cite{sun2023receptive}    & 2023 & IEEE TGRS      & 59.08 & 77.16   & 68.49 & 82.68   &  2.74 &  50.44 \\
       
       UIUNet \cite{wu2022uiu}          & 2022 & IEEE TIP       & 56.38 & 75.25   & 65.15 & 81.13   & 53.06 & 456.70 \\
       
       SANet \cite{zhu2023sanet}        & 2023 & ICASSP         & 64.54 & 80.49   & 62.17 & 78.64   & 12.04 &  42.04 \\
       
       MSHNet \cite{liu2024infrared}    & 2024 & CVPR           & 63.21 & 79.91   & 60.82 & 77.64   &  6.59 &  69.59 \\
       
       RPCANet \cite{wu2024rpcanet}     & 2024 & WACV           & 56.50 & 75.73   & 62.28 & 79.22   &  3.21 & 382.69 \\
       
       \textit{Ours}                    & 2025 & –             
         & \cellcolor{best}\textbf{89.90~\textcolor{delta_pos}{(+23.33)}} 
         & \cellcolor{best}\textbf{90.40~\textcolor{delta_pos}{(+8.32)}}
         & \cellcolor{best}\textbf{86.80~\textcolor{delta_pos}{(+16.34)}} 
         & \cellcolor{secondbest}83.26~\textcolor{delta_neg}{(-1.20)}
         & \cellcolor{best}\textbf{2.03~\textcolor{delta_pos}{(-0.71)}}
         & \cellcolor{best}\textbf{37.40~\textcolor{delta_pos}{(-4.64)}} \\
       \midrule
       \multicolumn{9}{l}{\textit{\textbf{Multi‐frame Methods}}} \\
       DTUM \cite{li2023direction}      & 2023 & IEEE TNNLS     & 71.48 & 85.26   & 67.97 & 82.79   &  9.64 & 128.16 \\
       
       TMP \cite{zhu2024tmp}            & 2024 & Expert Syst. Appl.& 70.03 & 83.97 & 77.73 & 88.67   & 16.41 &  92.85 \\
       
       ST-Trans \cite{tong2024st}       & 2024 & IEEE TGRS      & 70.04 & 83.91   & 76.02 & 87.50   & 38.13 & 145.16 \\
       
       SSTNet \cite{chen2024sstnet}     & 2024 & IEEE TGRS      & 71.55 & 85.11   & 76.96 & 88.07   & 11.95 & 123.59 \\

       Tridos \cite{duan2024triple}     & 2024 & IEEE TGRS      & 73.72 & 86.85   & 80.41 &  \cellcolor{best} \textbf{90.65}   & 14.13 & 130.72 \\

       STC \cite{zhu2025spatial}        & 2025 & Image Vis      & 74.03 & 86.87   & 80.71 & 90.42   & 10.75 &  59.58 \\
              
       \textit{Ours}                    & 2025 & –              
         & \cellcolor{best}\textbf{89.90~\textcolor{delta_pos}{(+15.87)}} 
         & \cellcolor{best}\textbf{90.40~\textcolor{delta_pos}{(+3.53)}}
         & \cellcolor{best}\textbf{86.80~\textcolor{delta_pos}{(+6.09)}} 
         & \cellcolor{secondbest}83.26~\textcolor{delta_neg}{(-7.39)}
         & \cellcolor{best}\textbf{2.03~\textcolor{delta_pos}{(-7.61)}}
         & \cellcolor{best}\textbf{37.40~\textcolor{delta_pos}{(-22.18)}} \\
       \bottomrule
     \end{tabular}
   }
  }
    \caption{\textbf{Multi-frame benchmark results}. Quantitative comparison of SIRST and MIRST methods on two sequence-based benchmarks (ITSDT-15k and IRDST). \colorbox{best}{\textbf{+}} denotes our model’s gain over the previous top baseline; \colorbox{secondbest}{\textbf{–}} denotes its shortfall.} 
\label{table1:sequence-based-dataset}
\end{table*}

\section{Quantitative Results}
\label{sec:results}

The quantitative results are presented in four parts. First, we benchmarked TY-RST on two multi-frame datasets (ITSDT-15k \cite{zhu2024tmp} and IRDST \cite{sun2023receptive}), comparing its performance against both SIRST and MIRST algorithms. To further highlight its effectiveness in diverse real-world scenarios, we additionally evaluated it on two single-frame datasets (NUAA-SIRST \cite{dai2021asymmetric} and NUDT-SIRST \cite{li2022dense}). Overall, TY-RST demonstrated clear superiority, outperforming even spatio-temporal models and achieving SOTA performance against $\textbf{20}$ different models across \textbf{four} benchmark datasets. Next, to assess its generalization capability, we conducted a cross-dataset validation experiment by training TY-RST on NUAA-SIRST and NUDT-SIRST and evaluating it on the unseen IRDST-1k benchmark \cite{zhu2024tmp}. Finally, we prepared two ablation studies on the ITSDT-15k benchmark for each component's effect and the best $C$ value in the NWD \cite{wang2021normalized} function.

\definecolor{benchmark}{HTML}{F0F8FF}     
\definecolor{best}{HTML}{E8FFE8}          
\definecolor{delta_pos}{HTML}{006400}     
\definecolor{delta_neg}{HTML}{F44336}     
\definecolor{secondbest}{HTML}{FFEFD5}   

\begin{table*}[h]
  \centering
  \Large
  {\renewcommand{\arraystretch}{1.4}%
   \setlength{\tabcolsep}{4pt}%
   \resizebox{\textwidth}{!}{%
     \begin{tabular}{c c c *{3}{c} *{3}{c} *{3}{c}}
       \toprule
       \cellcolor{benchmark}\textbf{Method}
         & \cellcolor{benchmark}\textbf{Year}
         & \cellcolor{benchmark}\textbf{Venue}
         & \multicolumn{3}{c}{\cellcolor{benchmark}\textbf{NUAA-SIRST (\%)}} 
         & \multicolumn{3}{c}{\cellcolor{benchmark}\textbf{NUDT-SIRST (\%)}} 
         & \multicolumn{3}{c}{\cellcolor{benchmark}\textbf{IRDST-1k (\%)}} 
         \\
       & & 
         & \textbf{$\mathrm{Pre}$} {\color{delta_pos}\(\uparrow\)} 
         & \textbf{$\mathrm{Rec}$} {\color{delta_pos}\(\uparrow\)}
         & \textbf{$\mathrm{F}_{1}$} {\color{delta_pos}\(\uparrow\)}
         & \textbf{$\mathrm{Pre}$} {\color{delta_pos}\(\uparrow\)} 
         & \textbf{$\mathrm{Rec}$} {\color{delta_pos}\(\uparrow\)} 
         & \textbf{$\mathrm{F}_{1}$} {\color{delta_pos}\(\uparrow\)}
         & \textbf{$\mathrm{Pre}$} {\color{delta_pos}\(\uparrow\)} 
         & \textbf{$\mathrm{Rec}$} {\color{delta_pos}\(\uparrow\)} 
         & \textbf{$\mathrm{F}_{1}$} {\color{delta_pos}\(\uparrow\)} 
         \\
         
       \cmidrule(lr){4-6}\cmidrule(lr){7-9}\cmidrule(lr){10-12}
       MDvsFA \cite{9009584} & 2018 & Remote Sens. & 84.5 & 50.7 & 59.7 & 60.8 & 19.2 & 26.2 &55.0&48.3&475.0\\
              
       ACLNet \cite{9570298} & 2021 & IEEE TGRS & 84.8 & 78.0 & 81.3 & 86.8 & 77.2 & 81.7 & 84.3&65.6&73.8 \\
       
       ACM \cite{dai2021asymmetric} & 2021 & WACV & 76.5 & 76.2 & 76.3 & 73.2 & 74.5 & 73.8 &67.9&60.5&64.0 \\

       ISNet \cite{zhang2022isnet} & 2022 & CVPR & 82.0 & 84.7 & 83.4 & 74.2 & 83.4 & 78.5 &71.8&74.1&72.9\\

       DNANet \cite{li2022dense} & 2023 & IEEE TIP & 84.7 & 83.6 & 84.1 & 91.4 & 88.9 & 90.1 & 76.8&72.1&74.4 \\
       
       AGPCNet \cite{zhang2023attention} & 2023 & IEEE TAES & 39.0 & 81.0 & 52.7 & 36.8 & 68.4 & 47.9 &41.5&47.0&44.1 \\
       
       EFLNet \cite{yang2024eflnet} & 2024 & IEEE TGRS & 88.2 & 85.8 & 87.0 & 96.3 & 93.1 & 94.7 & \cellcolor{best}\textbf{87.0} &	\cellcolor{best}\textbf{81.7} & \cellcolor{best}\textbf{84.3} \\
       \textit{Ours} & 2025 & – 
         & \cellcolor{best}\textbf{92.9~\textcolor{delta_pos}{(+4.7)}} & \cellcolor{best}\textbf{92.1~\textcolor{delta_pos}{(+6.3)}} & \cellcolor{best}\textbf{92.5~\textcolor{delta_pos}{(+5.5)}}
         & \cellcolor{best}\textbf{96.8~\textcolor{delta_pos}{(+0.5)}} & \cellcolor{best}\textbf{95.8~\textcolor{delta_pos}{(+2.7)}} & \cellcolor{best}\textbf{96.3~\textcolor{delta_pos}{(+1.6)}}
         & \cellcolor{secondbest}81.0\textbf{*} ~\textcolor{delta_neg}{(-6.0)} &  \cellcolor{secondbest}75.2\textbf{*} ~\textcolor{delta_neg}{(-6.5)}  & \cellcolor{secondbest}78.0\textbf{*} ~\textcolor{delta_neg}{(-6.3)} \\
       \bottomrule
     \end{tabular}
   }%
  }
\caption{\textbf{Single-frame benchmark results}. Quantitative comparison of SIRST algorithms on three single-frame benchmarks (IRDST-1k, NUAA-SIRST - using $2$ heads $P_2$ and $P_3$ -, and NUDT-SIRST). \textbf{*} denotes the \textbf{Cross-Dataset Validation} on IRDST-1k, where the model was trained on NUAA-SIRST and NUDT-SIRST and tested on IRDST-1k. Color codes match Table~\ref{table1:sequence-based-dataset}.}
\label{table2:single-frame-datasets}

\end{table*}


\subsection{Multi-Frame Benchmark Results}

In comparison with other SOTA algorithms, we primarily focused on learning-based methods due to their advanced and competitive performance. Since most of the SOTA SIRST algorithms are segmentation-based, we obtained their detection performance results from Chen et al. \cite{chen2024sstnet} and adopted the same data splits and image resolutions to ensure a fair comparison. Table~\ref{table1:sequence-based-dataset} summarizes the performance of our model compared to $10$ SIRST and $6$ MIRST algorithms on the ITSDT-15k and IRDST benchmarks. 

For the SIRST algorithms on the ITSDT-15k dataset, our model achieved the best results with mAP@$50$ of $86.80$\%, $16.34$\% higher than the second-best algorithm, DNANet \cite{li2022dense}. In terms of $F_{1}$ score, our model achieved the second-best result with a score of $83.26$\%, which is $1.20$ lower than DNANet's top score of $84.46$\%. However, our model is $\sim$$3.5$ times lighter in terms of the number of parameters ($2.03$M compared to $7.2$M) and $\sim$$3.6$ times less complex in terms of FLOPs ($37.40$ GFLOPs compared to $135.24$ GFLOPs) compared to DNANet.

For SIRST algorithms on the IRDST dataset, our model achieved the best results in both mAP@$50$ and $F_{1}$ score, outperforming the second-best model (RISTD \cite{hou2021ristdnet}) by $23.33$\% and $8.32$\%, respectively. Regarding model efficiency, our model also led with $0.71$M fewer parameters than the second-lightest model, RDIAN \cite{sun2023receptive}, and $4.64$ GFLOPS fewer than the second-fastest model, SANet \cite{zhu2023sanet}.

On the ITSDT-15k benchmark for MIRST algorithms, our model achieved the best mAP@$50$ score of $86.80$\%, outperforming the next best model, STC \cite{zhu2025spatial}, by $6.09$\%. In terms of $F_{1}$ score, our model scored $7.39$\% lower than the top-performing algorithm, Tridos \cite{duan2024triple}. However, our model is $\sim$$7$ times lighter in terms of the number of parameters ($2.03$M compared to $14.13$M) and runs $\sim$$1.6$ times faster ($37.40$ GFLOPS compared to $59.58$ GFLOPS).

On the IRDST benchmark for MIRST algorithms, our model again achieved the best performance in both mAP@$50$ and $F_{1}$ score, outperforming the next best model by $15.87$\% and $3.53$\%, respectively. Regarding efficiency, our model set a best result with $7.61$M fewer parameters than the second-lightest model, DTUM \cite{li2023direction}, and $22.18$ GFLOPS fewer than the second-fastest model, STC \cite{zhu2025spatial}. 

Finally, to test our model’s real-time performance capabilities, we ran it on a single NVIDIA RTX $3080$ Ti laptop GPU. On the ITSDT-15k benchmark, using $P_2$ head only and by removing the entire PAN network, our model achieved $\sim$123 FPS.


\subsection{Single-Frame Benchmark Results}

For further validation of our model’s effectiveness, we benchmarked it on two single-frame datasets (NUAA-SIRST \cite{dai2021asymmetric} and NUDT-SIRST \cite{li2022dense}), which feature diverse and complex real-life and synthetic challenges with varied backgrounds such as sea, buildings, and urban scenes. Since most of the selected SIRST algorithms in this experiment are segmentation-based, we used detection performance results from Yang et al. \cite{yang2024eflnet} work and adopted the same data splits and image resolutions to ensure fair comparison. It is worth noting that this line of work \cite{yang2024eflnet} did not report mAP results; therefore, we excluded mAP from our evaluation. Based on Table~\ref {table2:single-frame-datasets}, our model achieved the best results across three evaluation metrics: precision, recall, and $F_{1}$ score, using a single $P_2$ Head for NUDT-SIRST benchmark, and two heads ($P_2$ and $P_3$) for NUAA-SIRST benchmark. Finally, in terms of the FPS, our model achieved $\sim$105 FPS due to the utilization of $P_2$ and $P_3$ heads, and a portion of the PAN network.


\definecolor{benchmark}{HTML}{F0F8FF}     
\definecolor{best}{HTML}{E8FFE8}          
\definecolor{delta_pos}{HTML}{006400}     
\definecolor{delta_neg}{HTML}{F44336}     
\definecolor{secondbest}{HTML}{FFEFD5}   

\renewcommand{\arraystretch}{1.5}
\begin{table*}[htbp]
\centering
\resizebox{\textwidth}{!}{%
\begin{tabular}{
>{\columncolor[HTML]{FFFFFF}}c 
>{\columncolor[HTML]{FFFFFF}}c 
>{\columncolor[HTML]{FFFFFF}}c 
>{\columncolor[HTML]{FFFFFF}}c 
>{\columncolor[HTML]{FFFFFF}}c 
>{\columncolor[HTML]{FFFFFF}}c 
>{\columncolor[HTML]{FFFFFF}}c 
>{\columncolor[HTML]{FFFFFF}}c 
>{\columncolor[HTML]{FFFFFF}}c 
>{\columncolor[HTML]{FFFFFF}}c 
>{\columncolor[HTML]{FFFFFF}}c 
>{\columncolor[HTML]{FFFFFF}}c 
>{\columncolor[HTML]{FFFFFF}}c}
\hline
\cellcolor{benchmark} & \cellcolor{benchmark} & \cellcolor{benchmark} & \cellcolor{benchmark} & \cellcolor{benchmark} & \multicolumn{2}{c}{\cellcolor{benchmark} \textbf{Model Trimming}} & \cellcolor{benchmark} & \cellcolor{benchmark} & \cellcolor{benchmark} & \cellcolor{benchmark} & \cellcolor{benchmark} & \cellcolor{benchmark} \\
\cline{6-7}
\multirow{-2}{*}{\cellcolor{benchmark}\textbf{YOLOv12n}} & 
\multirow{-2}{*}{\cellcolor{benchmark} \textbf{Stride}} & 
\multirow{-2}{*}{\cellcolor{benchmark} \textbf{NWD}} & 
\multirow{-2}{*}{\cellcolor{benchmark} \textbf{$\textbf{C}_{\textbf{2}}$ + $\textbf{P}_{\textbf{2}}$ Head}} & 
\multirow{-2}{*}{\cellcolor{benchmark} \textbf{Coord. Attn.}} & 
\multicolumn{1}{c}{\cellcolor{benchmark} \textbf{- $\textbf{P}_{\textbf{3}}$$\textbf{P}_{\textbf{4}}$$\textbf{P}_{\textbf{5}}$}} & 
\cellcolor{benchmark} \textbf{- PAN}  & 
\multirow{-2}{*}{\cellcolor{benchmark}\textbf{{ \textbf{(\%)} $\mathrm{\textbf{mAP}}_{\textbf{50}}$ } {\color{delta_pos}\(\uparrow\)}}} & 
\multirow{-2}{*}{\cellcolor{benchmark}\textbf{{\textbf{(\%)} $\mathrm{\textbf{Pre}}$} {\color{delta_pos}\(\uparrow\)}}} & 
\multirow{-2}{*}{\cellcolor{benchmark}\textbf{{\textbf{(\%)} $\mathrm{\textbf{Rec}}$} {\color{delta_pos}\(\uparrow\)}}} & 
\multirow{-2}{*}{\cellcolor{benchmark}\textbf{{\textbf{(\%)} $\mathrm{\textbf{F}}_{\textbf{1}}$} {\color{delta_pos}\(\uparrow\)}}} & 
\multirow{-2}{*}{\cellcolor{benchmark}\begin{tabular}[c]{@{}c@{}} \textbf{Params (M)} {\color{delta_pos}\(\downarrow\)} \end{tabular}} & 
\multirow{-2}{*}{\cellcolor{benchmark}\begin{tabular}[c]{@{}c@{}} \textbf{FLOPS (G)} {\color{delta_pos}\(\downarrow\)} \end{tabular}} \\
\ding{51} & \ding{55} & \ding{55} & \ding{55} & \ding{55} & \multicolumn{1}{c}{\ding{55}} & \ding{55} & \textbf{78.9} & \textbf{85.0} & \textbf{68.8} & \textbf{76.05} & \textbf{2.56} & \textbf{6.3} \\
\ding{51} & \ding{51} & \ding{55} & \ding{55} & \ding{55} & \multicolumn{1}{c}{\ding{55}} & \ding{55} & 84.1 & 86.9 & 75.8 & 80.97 & 2.56 & 25.3 \\
\ding{51} & \ding{51} & \ding{51} & \ding{55} & \ding{55} & \multicolumn{1}{c}{\ding{55}} & \ding{55} & 85.1 & 85.3 & 79.7 & 82.40 & 2.56 & 25.3 \\
\ding{51} & \ding{51} & \ding{51} & \ding{51} & \ding{55} & \multicolumn{1}{c}{\ding{55}} & \ding{55} & 86.3 & 88.4 & 78.5 & 83.16 & 2.72 & 50.0 \\
\ding{51} & \ding{51} & \ding{51} & \ding{51} & \ding{51} & \multicolumn{1}{c}{\ding{55}} & \ding{55} & 86.5 & 87.6 & 79.2 & 83.19 & 2.73 & 50.2 \\
\ding{51} & \ding{51} & \ding{51} & \ding{51} & \ding{51} & \multicolumn{1}{c}{\ding{51}} & \ding{55} &  86.8 & 88.0 & 79.0 & 83.26 & 2.73 & 43.4 \\
\textbf{\ding{51}} & \textbf{\ding{51}} & \textbf{\ding{51}} & \textbf{\ding{51}} & \textbf{\ding{51}} & \multicolumn{1}{c}{\textbf{\ding{51}}} & \textbf{\ding{51}} & \cellcolor{best}\textbf{86.8~\textcolor{delta_pos}{(+7.9)}} &  \cellcolor{best}\textbf{88.0~\textcolor{delta_pos}{(+3.0)}} &  \cellcolor{best}\textbf{79~\textcolor{delta_pos}{(+10.2)}} & \cellcolor{best}\textbf{83.26~\textcolor{delta_pos}{(+7.21)}} & \cellcolor{best}\textbf{2.03~\textcolor{delta_pos}{(-0.53)}} &  \cellcolor{secondbest}37.40~\textcolor{delta_neg}{(+31.1)}\\
\hline
\ding{51}&\ding{51}& \ding{51}&\ding{51} &\ding{51} & \ding{51} & \ding{51} &- & 92.2 & 83.2 &87.47 & 2.03 & 37.40\\
\ding{51}&\ding{51}& \ding{51}&\ding{51} &\ding{51} & $\textbf{P}_{\textbf{2}}$$\textbf{P}_{\textbf{3}}$ & \textbf{- Partial PAN} &- & \textbf{92.9} & \textbf{92.1} & \textbf{92.5} & \textbf{2.10} & \textbf{40.30}\\
\hline
\end{tabular} }
\caption{Ablation Study performed on ITSDT-15k Benchmark (\textbf{upper part}) and NUAA-SIRST Benchmark (\textbf{lower part}). \colorbox{best}{\textbf{+}} denotes our model’s gain over the baseline YOLOv12n model; \colorbox{secondbest}{\textbf{–}} denotes its shortfall.}
\label{table:main-ablation}
\end{table*}
\renewcommand{\arraystretch}{1.0}  

\subsection{Cross-Dataset Validation Results}

In this experiment, we aimed to assess the generalization capability of our model on an unseen dataset. We carefully chose to train our model on the NUAA-SIRST and NUDT-SIRST datasets and evaluate it on the unseen IRDST-1k dataset \cite{zhu2024tmp}, as the background characteristics of the training datasets closely resemble those of the validation set. In other words, all three datasets share similarly challenging scenarios with complex backgrounds, including clouds, sea, buildings, and fields. However, the specific instance images differ, ensuring that this experiment evaluates the model’s ability to generalize to new data from a similar distribution. The results are presented in Table~\ref{table2:single-frame-datasets}, where our model is marked with $\textbf{*}$ to indicate cross-dataset validation on IRDST-1k. Our model demonstrated strong generalization capability, ranking second in terms of $F_{1}$ score and Recall, and third in Precision, outperforming up to six algorithms that were trained on the IRDST-1k benchmark.


\subsection{Ablation Study}

The ablation study in this work consists of three main parts. First, we conduct a detailed analysis of the impact of each experiment performed on the baseline YOLOv12n model to arrive at the final version of our proposed TY-RIST model. Second, we present a tuning study of the $C$ parameter embedded in the NWD regression loss function. Finally, we present a case study that replicates the experiments on YOLOv12s~\cite{tian2025yolov12}.

\subsubsection{Impact of Each Component}

To highlight the impact of each of the six experiments on our model’s performance, we conducted a comprehensive ablation study on the ITSDT-15k benchmark presented in the upper part of Table~\ref{table:main-ablation}. The first experiment involved evaluating the vanilla YOLOv12n model on the ITSDT-15k benchmark, achieving a mAP@$50$ of 78.9\%, which indicates the strong baseline performance of the chosen model. Reducing the stride improved mAP@$50$ by $5.2$\%, indicating the extraction of higher-quality features and thus solving the minimal feature challenge, but resulted in an additional computational cost of 19 GFLOPS. Replacing CIoU with the NWD function boosted mAP@$50$ by $1$\% due to solving the instability challenge in the CIoU function, without increasing computational cost. Adding the higher resolution feature map $C_{2}$ and its corresponding detection head $P_{2}$ further boosted the Precision by $3.1$\%, thus reducing the false alarm rate, but $\sim$doubled the GFLOPS. Integrating the CA blocks on the $P_{2}$ head resulted in a  $0.7$\% increase in Recall, thus reducing the missed detection rate, but resulted in a minimal $0.2$ GFLOPS increase in computational cost. Removing the $P_{3}$, $P_{4}$, and $P_{5}$ detection heads improved mAP@$50$ by 0.3\% and, more importantly, reduced computational cost by $6.8$ GFLOPS, while deactivating the PAN network maintains performance while reducing the number of parameters by $0.53$M and GFLOPS by $6$. Overall, the model trimming experiment reduced the GFLOPS by $\sim$25.5\% and the number of parameters by $\sim$25.6\%. In the lower part of Table~\ref{table:main-ablation}, the ablation study on the number of heads used for the NUAA-SIRST dataset is presented. Adding extra head and parts of the PAN network increased the number of parameters by $0.07$M and GFLOPS by $2.9$. In addition, the introduction of the second head improved the recall by $8.9$\% and precision by $0.7$\%.

\subsubsection{Fine-tuning the $C$ Parameter in NWD}

As mentioned earlier in Section~\ref{metrics}, the NWD function includes a tunable, per-dataset parameter $C$. Table~\ref{tab:nwd} summarizes the ablation study conducted to select the optimal value of $C$ from the set \{$9, 11, 13, 15, 17$\} for the ITSDT-15k benchmark, with $17$ identified as the best-performing value. These tested values were inspired by the work of \cite{yang2024eflnet}. A critical observation from this study is that some values of $C$ may lead to a performance drop compared to the vanilla CIoU loss function.
 
\definecolor{benchmark}{HTML}{F0F8FF}     
\definecolor{best}{HTML}{E8FFE8}          
\definecolor{delta_pos}{HTML}{006400}     
\definecolor{delta_neg}{HTML}{F44336}     
\definecolor{delta_pos1}{HTML}{000000}     

\definecolor{benchmark}{HTML}{F0F8FF}
\renewcommand{\arraystretch}{1.2}
\begin{table}[]
\centering
\resizebox{\columnwidth}{!}{%
\begin{tabular}{
>{\columncolor[HTML]{FFFFFF}}c 
>{\columncolor[HTML]{FFFFFF}}c 
>{\columncolor[HTML]{FFFFFF}}c 
>{\columncolor[HTML]{FFFFFF}}c 
>{\columncolor[HTML]{FFFFFF}}c }
\hline
\cellcolor{benchmark} {$\mathrm{\textbf{C}}$}       &  \cellcolor{benchmark}\textbf{{\textbf{(\%)} $\mathrm{\textbf{mAP}}_{\textbf{50}}$} \cellcolor{benchmark}  {\color{delta_pos1}\(\uparrow\)}} &  \cellcolor{benchmark} \textbf{{ \textbf{(\%)} $\mathrm{\textbf{Pre}}$}  {\color{delta_pos1}\(\uparrow\)}}  & \cellcolor{benchmark} \textbf{{\textbf{(\%)} $\mathrm{\textbf{Rec}}$}  {\color{delta_pos1}\(\uparrow\)}}  &  \cellcolor{benchmark}\textbf{{\textbf{(\%)} $\mathrm{\textbf{F}}_{\textbf{1}}$} {\color{delta_pos1}\(\uparrow\)}}    \\ \hline

\textbf{baseline+Stride} & \textbf{84.1}            & \textbf{86.9}          & \textbf{75.8}          & \textbf{80.97}          \\

9               & 84.5            & 83.3          & 77.6          & 80.35          \\ 
11              & 84.4            & 84.5          & 76.3          & 80.19          \\ 
13              & 84.1            & 83.1          & 78.4          & 80.68          \\ 
15              & 83.4            & 81.6          & 77.1          & 79.29          \\ 
\textbf{17}     &   \cellcolor{best}\textbf{85.1~\textcolor{delta_pos}{(+1.0)}}   &   \colorbox{secondbest}{85.3~\textcolor{delta_neg}{(-1.6)}} &  \cellcolor{best}\textbf{79.7 ~\textcolor{delta_pos}{(+3.9)}} &  \cellcolor{best}\textbf{82.40 ~\textcolor{delta_pos}{(+1.43)}}\\ \hline
\end{tabular}}
\caption{Ablation Study on NWD with Different C Values. \colorbox{best}{\textbf{+}} denotes our model’s gain over the \textbf{baseline + stride} model; \colorbox{secondbest}{\textbf{–}} denotes its shortfall.}
\label{tab:nwd}
\end{table}
\renewcommand{\arraystretch}{1.0}  

\subsubsection{Replicating Experiments on YOLOv12s}

Table~\ref{tab:yolov12s} summarizes the replicated experiments on YOLOv12s~\cite{tian2025yolov12}, further demonstrating their generalizability across different YOLO models.


\definecolor{benchmark}{HTML}{F0F8FF}     
\definecolor{best}{HTML}{E8FFE8}          
\definecolor{delta_pos}{HTML}{006400}     
\definecolor{delta_neg}{HTML}{F44336}     
\definecolor{delta_pos1}{HTML}{000000}     

\definecolor{benchmark}{HTML}{F0F8FF}
\renewcommand{\arraystretch}{1.2}
\begin{table}[h]
\centering
\resizebox{\columnwidth}{!}{%
\begin{tabular}{
>{\columncolor[HTML]{FFFFFF}}c 
>{\columncolor[HTML]{FFFFFF}}c 
>{\columncolor[HTML]{FFFFFF}}c 
>{\columncolor[HTML]{FFFFFF}}c
>{\columncolor[HTML]{FFFFFF}}c}
\hline
\cellcolor{benchmark} {$\mathrm{\textbf{Experiment Details}}$}       &  \cellcolor{benchmark}\textbf{{\textbf{(\%)} $\mathrm{\textbf{mAP}}_{\textbf{50}}$} \cellcolor{benchmark}  {\color{delta_pos1}\(\uparrow\)}} &   \cellcolor{benchmark}\textbf{{\textbf{(\%)} $\mathrm{\textbf{F}}_{\textbf{1}}$} {\color{delta_pos1}\(\uparrow\)}}    & 
\cellcolor{benchmark}\textbf{{ $\mathrm{\textbf{Params (M)}}$} {\color{delta_pos1}\(\downarrow\)}} &
\cellcolor{benchmark}\textbf{{ $\mathrm{\textbf{FLOPS (G)}}$} {\color{delta_pos1}\(\downarrow\)}}
\\ \hline

YOLOv12s \cite{tian2025yolov12} & 81.80 & 76.87 &9.23 &21.20 \\
+ Stride Exp.& 85.40& 79.89 & 9.23& 84.8\\ 
+ NWD Exp. (C=15)& 86.10 & 81.23 & 9.23& 84.8  \\
+ P2 Head Exp. & 86.60& 83.83&9.69&143.10  \\ 
+ CA Exp. (3 block)& 87.60& 84.00& 9.70&143.30        \\ 
+ P$_2$ Head Only - PAN Exp.& 
\cellcolor{best}\textbf{87.81~\textcolor{delta_pos}{(+6.01)}} & \cellcolor{best}\textbf{84.24~\textcolor{delta_pos}{(+7.37)}}  & \cellcolor{best}\textbf{6.92~\textcolor{delta_pos}{(-2.78)}} & \cellcolor{secondbest}131.50 ~\textcolor{delta_neg}{(+110.30)}  \\ \hline
\end{tabular}}
\caption{Replicating the Experiments on YOLOv12s on ITSDT-15k Benchmark. \colorbox{best}{\textbf{+}} denotes our model’s gain over the baseline YOLOv12s model; \colorbox{secondbest}{\textbf{–}} denotes its shortfall.}
\label{tab:yolov12s}
\end{table}
\renewcommand{\arraystretch}{1.0}  



\section{Conclusion}
\label{sec:conc}

This work proposes TY-RIST, an efficient real-time infrared small target detection algorithm based on the latest YOLO family member, YOLOv12n. With a series of experiments, TY-RIST achieved SOTA against $20$ different models. Nonetheless, there remains room for improvement, particularly in further reducing false alarms and missed detections by integrating temporal features and effectively fusing them with spatial features—an area reserved for future work.

\section{Acknowledgments}
\label{ack}
We thank Prof. Jiaming Zheng from Hunan University (InSAI Lab) for his collaboration. This work was carried out on the bwForCluster Helix, supported by the state of Baden-Württemberg through bwHPC and the German Research Foundation.

{
    \small
    \bibliographystyle{ieeenat_fullname}
    \bibliography{main}
}

\end{document}